\documentclass[11pt,letterpaper]{article} 
\usepackage[parfill]{parskip} 
\usepackage{mathtools}
\usepackage{xfrac}
\usepackage{dsfont}
\usepackage[top=0.85in,left=1in,footskip=0.75in]{geometry}
\usepackage{algorithm} 
\usepackage[noend]{algpseudocode} 
\usepackage{array} 
\usepackage{float} 
\usepackage{amssymb} 
\usepackage{color} 
\usepackage{soul}
\usepackage[bordercolor=white,backgroundcolor=gray!30,linecolor=black, colorinlistoftodos]{todonotes}

\global\long\def\iid{\overset{i.i.d.}{\sim}}
\newcommand{\K}{K_{\xb\xb}}
\newcommand{\xb}{\mathbf{x}}

\usepackage[english]{babel}
 \usepackage{amsthm}

\usepackage{graphicx, caption, subcaption}

\newtheorem{theorem}{Theorem}
\makeatother

\usepackage{changepage}

\usepackage[utf8]{inputenc}

\usepackage{textcomp,marvosym}

\usepackage{amsmath,amssymb}

\usepackage{cite}

\usepackage{nameref,hyperref}

\usepackage[right]{lineno}

\usepackage{microtype} \DisableLigatures[f]{encoding = *, family = * }

\usepackage{rotating}

\usepackage[aboveskip=1pt,labelfont=bf,labelsep=period,justification= raggedright,singlelinecheck=off]{caption}

\makeatletter \renewcommand{\@biblabel}[1]{\quad#1.} \makeatother

\date{}

\usepackage{lastpage,fancyhdr,graphicx} 
\usepackage{epstopdf}
 \fancyheadoffset[L]{0in} \fancyfootoffset[L]{0in}  \usepackage{authblk}

\title{\LARGE \bf
Spatial Mapping with Gaussian Processes and Nonstationary Fourier Features}

\author[1]{Jean-Francois Ton} \author[2]{Seth Flaxman} \author[1]{Dino Sejdinovic}  \author[3,*]{Samir Bhatt}

\affil[1]{Department of Statistics, University of Oxford, Oxford, OX1 3LB, UK}
\affil[2]{Department of Mathematics and Data Science Institute, Imperial College London, London, SW7 2AZ, UK}
\affil[3]{Department of Infectious Disease Epidemiology, Imperial College London, London, W2 1PG, UK}%
\affil[*]{Corresponding author: bhattsamir@gmail.com}

\begin{document}

\maketitle
\thispagestyle{empty}
\pagestyle{empty}

\begin{abstract}

The use of covariance kernels is ubiquitous in the field of spatial statistics. Kernels allow data to be mapped into high-dimensional feature spaces and can thus extend simple linear additive methods to nonlinear methods with higher order interactions. However, until recently, there has been a strong reliance on a limited class of stationary kernels such as the Mat\'ern or squared exponential, limiting the expressiveness of these modelling approaches. Recent machine learning research has focused on spectral representations to model arbitrary stationary kernels and introduced more general representations that include classes of nonstationary kernels. In this paper, we exploit the connections between Fourier feature representations, Gaussian processes and neural networks to generalise previous approaches and develop a simple and efficient framework to learn arbitrarily complex nonstationary kernel functions directly from the data, while taking care to avoid overfitting using state-of-the-art methods from deep learning. We highlight the very broad array of kernel classes that could be created within this framework.  We apply this to a time series dataset and a remote sensing problem involving land surface temperature in Eastern Africa. We show that without increasing the computational or storage complexity, nonstationary kernels can be used to improve generalisation performance and provide more interpretable results.

\end{abstract}

\section{Introduction}

The past decade has seen a tremendous and ubiquitous growth in both data collection and computational resources available for data analysis. In particular, spatiotemporal modelling has been brought to the forefront with applications in finance \cite{2000APrices}, weather forecasting\cite{Berrocal2007CombiningForecasts}, remote sensing\cite{2002SpatialSensing,Weiss2014AirPrediction,Weiss2014AnTime-series} and demographic/disease mapping\cite{Bhatt2015The2015,Bhatt2013TheDengue,hay2013global}. The methodological workhorse of mapping efforts has been Gaussian process (GP) regression \cite{Rasmussen2006GaussianLearning,Diggle2007Model-basedGeostatistics}. The reliance on GPs stems from their convenient mathematical framework which allows the modelling of distributions over nonlinear functions. GPs offer robustness to overfitting, a principled way to integrate over hyperparameters, and provide uncertainty intervals. In a GP, every point in some continuous input space is associated with a normally distributed random variable, such that every finite collection of those random variables has a multivariate normal distribution -- entirely defined through a mean function $\mu(\cdot)$ and a covariance kernel function $k(\cdot,\cdot)$. In many settings, $\mu(\cdot)=0$ and modelling proceeds through selecting the appropriate kernel function which entirely determines the properties of the GP, and can have a significant influence on both the predictive performance and on the model interpretability\cite{Paciorek2006SpatialFunctions,genton2001classes}. However, in practice, the kernel function is often (somewhat arbitrarily) set \emph{a priori} to the squared exponential or Mat\'ern class of kernels \cite{genton2001classes}, justifying this choice by the fact that they model a rich class of functions \cite{Micchelli2006}. 

While offering an elegant mathematical framework, performing inference with GP models is computationally demanding. Namely, evaluating the GP posterior involves a matrix inversion, which for $n$ observations, requires $\mathcal{O}(n^3)$ time and $\mathcal{O}(n^2)$ storage complexity. This makes fitting full GP models prohibitive for any dataset that exceeds a few thousand observations (thereby limiting their use exactly in the regimes where a flexible nonlinear model is of interest). In response to these limitations, many scalable approximations to GP modelling have been proposed. 
Examples include inducing points representations \cite{quinonero2005unifying}, the Nystr\"{o}m approximation \cite{Rasmussen2006GaussianLearning}, the Fully Independent Training Conditional (FITC) model \cite{Snelson2012VariableProcesses}, stochastic partial differential equation representations \cite{Lindgren2011AnApproach}, and representations in the Fourier domain \cite{Rahimi2008RandomMachines,Lazaro-Gredilla2010SparseRegression,Samo2015GeneralizedKernels2}. Most of these approaches reduce the computational complexity to $\mathcal{O}(nm^2)$ time and $\mathcal{O}(nm)$ storage for $m\ll n$ playing the role of the number of dimensions of the functional space under consideration (e.g. the number of inducing points or the number of frequencies in Fourier representations).

In this contribution, we will focus on large-scale Fourier representations of GPs. These methods traditionally rely on the strong assumption of the stationarity (or shift-invariance) of kernel functions, which is made in the vast majority of applications (and is indeed satisfied by the most often used squared exponential and Mat\'ern kernels). Stationarity in the spatio-temporal data means that the similarity between two responses in space and time does not depend on the location and time itself, but only on the difference (or lag) between them, i.e. kernel function can be written as $k(x_1,x_2)=\kappa(x_1-x_2)$ for some function $\kappa$. Several recent works \cite{Lazaro-Gredilla2010SparseRegression,wilson2013gaussian,Yang2015AKernels} consider flexible families of kernels based on Fourier representations, thus avoiding the need to choose a specific kernel a priori and allowing the kernel to be learned from the data, but these approaches are restricted to the stationary case. In many applications, particularly when data is rich, relaxing the assumption of stationarity can greatly improve generalisation performance \cite{Paciorek2006SpatialFunctions}. To address  this, recent work in \cite{Samo2015GeneralizedKernels2,genton2001classes} note that a more general spectral characterisation exists that includes nonstationary kernels \cite{Yaglom1987CorrelationFunctions,genton2001classes} and uses it to construct  nonstationary kernel families. In this paper, we build on the work of \cite{Samo2015GeneralizedKernels2,Lazaro-Gredilla2010SparseRegression,Remes2017Non-StationaryKernels} and develop a simple and practicable framework for learning spatiotemporal nonstationary kernel functions directly from the data by exploiting the connections between Fourier feature representations, Gaussian processes and neural networks\cite{Rasmussen2006GaussianLearning}. Specifically, we directly learn frequencies in nonstationary spectral kernel representations using an appropriate neural network architecture, and adopt techniques used for deep learning regularisation \cite{Srivastava2014} to prevent overfitting. We demonstrate the utility of the proposed method for learning nonstationary kernel functions in a time series example and in spatial mapping of land surface temperature in East Africa.

\section{Methods and Theory}

\subsection{Gaussian Process Regression}

Gaussian process regression (GPR) takes a training dataset $\mathcal{D}=\{(x_i,y_i)\}_{i=1}^n$ where $y_i\in \mathbb R$ is real-valued response/output and $x_i\in \mathbb R^D$ is a $D$-dimensional input vector. The response $y_i$ and the input $x_i$ are connected via the observation model 
\begin{equation}
\label{eq:normal_obs}
y_i = f(x_i)+\epsilon_i,\quad \epsilon_i \iid \mathcal{N}(0,\sigma_n^2),\quad i=1,\ldots,n.
\end{equation}
GPR is a Bayesian non-parametric approach that imposes a prior distribution on functions $f$, namely a GP prior, such that any vector ${\bf f}= \left[f(x_1),..,f(x_m)\right]$ of a finite number of evaluations of $f$ follows a multivariate normal distribution ${\bf f} \sim \mathcal{N}(0,\K)$, where the covariance matrix $\K$ is created as a Gram matrix based on the kernel function evaluations, $[K_{\xb\xb}]_{ij}=k(x_i,x_j)$. Throughout this paper we will assume that the mean function of the GP prior is $\mu=0$, however, all the approaches in this paper can be easily extended to include a mean function\cite{Bhatt2017}. In stationary settings, $k(x_i,x_j)=\kappa(x_i-x_j)$ for some function $\kappa(\delta)$. A popular choice is the automatic relevance determination (ARD) kernel\cite{Rasmussen2006GaussianLearning}, given by $\kappa(\delta)=\tau^2\exp(-\delta^\top\Lambda\delta)$ where $\tau^2>0$ and $\Lambda=\text{diag}(\lambda_1,\ldots,\lambda_D)$. Kernel $k$ will typically have hyperparameters $\theta$ (e.g. $\theta=[\tau,\lambda_1,\ldots,\lambda_D]$ for the ARD kernel) and one can thus consider a Bayesian hierarchical model:
\begin{eqnarray} 
\theta & \sim & \pi(\theta)\nonumber\\ 
f |\theta & \sim & GP(0,k_{\theta})\nonumber \\ 
y_i|f,x_i,\theta & \sim & \mathcal N(f(x_i),\sigma_n^2),\quad i=1,\ldots,n. 
\end{eqnarray} 

The posterior predictive distribution is straightforward to obtain from the conditioning properties of multivariate normal distributions. For a new input $x^*$, we can find the posterior predictive distribution of the associated response $y^*$
\begin{eqnarray}
p(y^*|x^*,\mathcal{D},\theta) &=& \mathcal{N}(y^*; \mu_\theta,\sigma_\theta^{2})\\
\mu_\theta &=&k_{x^*\xb}(\K+\sigma_n^2 I_n)^{-1}y\\
\sigma_\theta^2 &=& \sigma_n^2 + k(x^*,x^*) - k_{x^*\xb}(\K+\sigma_n^2 I_n)^{-1} k_{\xb x^*},
\end{eqnarray}
where $k_{\xb x^*}=[k(x_1,x^*),\ldots,k(x_n,x^*)]^\top$, $k_{x^*\xb}=k_{\xb x^*}^\top$ and it is understood that the dependence on $\theta$ is through the kernel $k=k_\theta$.
The computational complexity in prediction stems from the matrix inversion $(\K+\sigma_n^2I_n)^{-1}$. The marginal likelihood (also called model evidence) of the vector of outputs ${\bf y}=[y_1,\ldots,y_n]$, is given by $p({\bf y}|\theta)=\int p({\bf y}|{\bf f},\theta)p({\bf f}|\theta)d{\bf f}$, is obtained by integrating out the GP evaluations $\bf f$ from the likelihood of the observation model. Maximising the marginal likelihood over hyperparameters allows for automatic regularisation and hence for selecting an appropriate model complexity. For a normal observation model in \eqref{eq:normal_obs}, the log marginal likelihood is available in closed form
\begin{equation}\label{eq:marginallikelihood}
\log p({\bf y}|\theta) = -\frac{n}{2}\log(2\pi) -\frac{1}{2}|\K+\sigma_n^2 I_n| - \frac{1}{2}{\bf y}^T\left(\K+\sigma_n^2 I_n\right)^{-1}{\bf y}.
\end{equation}
Computing the inverse and determinant in \eqref{eq:marginallikelihood} are computationally demanding - moreover, they need to be computed for every hyperparameter value $\theta$ under consideration. To allow for computational tractability, we will use an approximation of $\K$ based on Fourier features (see section 2.3 and 2.4). 

Alternative representations can easily be used such as the primal/dual representations in a closely related frequentist method, kernel ridge regression (KRR) \cite{Hastie2009TheLearning}. In contrast to KRR, optimising the marginal likelihood as above retains the same computational complexity while providing uncertainty bounds and automatic regularisation without having to tune a regularisation hyperparameter. However, the maximisation problem of \eqref{eq:marginallikelihood} is non-convex thereby limiting the chance of finding a global optimum, but instead relying on  reasonable local optima \cite{Rasmussen2006GaussianLearning}.

\subsection{Random Fourier Feature mappings}
The Wiener-Khintchine theorem states that the power spectrum and the autocorrelation function of a random process constitute a Fourier pair. Given this, random Fourier feature mappings and similar methodologies\cite{Remes2017Non-StationaryKernels,Yang2015AKernels,Samo2015GeneralizedKernels2,Lazaro-Gredilla2010SparseRegression,Rahimi2008RandomMachines} appeal to Bochner's theorem to reformulate the kernel function in terms of its spectral density.
\begin{theorem}
(Bochner's Theorem) A stationary continuous kernel $k(x_i,x_j)=\kappa(x_i-x_j)$ on $\mathbb{R}^d$ is positive definite if and only if $\kappa(\delta)$ is the Fourier transform of a non-negative measure.

\end{theorem}
Hence, for an appropriately scaled shift invariant complex kernel $\kappa(\delta)$, i.e. for $\kappa(0)=1$, Bochner's Theorem ensures that its inverse Fourier Transform is a probability measure:
\begin{equation} \label{bochners}
k(x_1,x_2) = \int_{\mathbb{R}^d}e^{i \omega^T (x_1-x_2)} \mathbb{P}(d\omega).
\end{equation}
Thus, Bochner's Theorem introduces the duality between stationary kernels and the spectral measures $\mathbb{P}(d\omega)$. Note that the scale parameter of the kernel, i.e. $\sigma_f^2=\kappa(0)$ can be trivially added back into the kernel construction by rescaling. Table \ref{tab1} shows some popular kernel functions and their respective spectral densities.

\begin{table}[H]
	\centering
	\begin{tabular}{l|ll}
		Kernel Name& $k(\delta)$ & $p(\omega)$ \\ \hline
		\small Squared exponential& $e^{-\frac{(\|\delta\|_2^2)}{2\sigma}}$ , $\sigma>0$    &     $(2\pi)^{-\frac{D}{2}}\sigma^D \exp(-\frac{\sigma^2\|\omega\|^2_2}{2})$        \\ \\
		\small Laplacian&   $\exp(-\sigma\|\delta\|_1)$, $\sigma>0$           &  
		$\left(\frac{2}{\pi}\right)^{\frac{2}{D}}\prod_{i=1}^{D}\frac{\sigma}{\sigma^2+\omega_i^2}$  \\ \\
		\small Mat\'ern&
		$\frac{2^{1-\lambda}}{\Gamma(\lambda)}\left(\frac{\sqrt{(2\lambda)\|\delta\|_2}}{\sigma}\right)^{\lambda}K_{\lambda}\left(\frac{\sqrt{(2\lambda)\|\delta\|_2}}{\sigma}\right)$ &
		$\frac{2^{D+\lambda}\pi^{\frac{D}{2}}\Gamma(\lambda+D/2)\lambda^{\lambda}}{\Gamma(\lambda)\sigma^{2\lambda}} \left(\frac{2\lambda}{\sigma^2}+4\pi^2\|\omega\|^2_2\right)^{-(\lambda+D/2)}$
		 \\ &
		 $\lambda>0, \sigma>0$
	\end{tabular}
	\caption{Summary table of kernels and their spectral densities}
    \label{tab1}
\end{table}

By taking the real part of equation \ref{bochners} (since we are commonly interested only in real-valued kernels in the context of GP modelling) and performing standard Monte Carlo integration, we can derive to a finite-dimensional, reduced rank approximation of the kernel function 
\begin{align}
\boldmath{k}(x_1,x_2)
&=\int_{\mathbb{R}^D}  e ^{i \omega^T (x_1-x_2)} \mathbb{P}(d\omega) \\
&= \mathbb{E}_{\omega\sim\mathbb{P}}\left[e^{i \omega^T (x_1-x_2)}\right], \\
&= \mathbb{E}_{\omega\sim\mathbb{P}}\left[  \cos(\omega^T (x_1-x_2)) + i  \sin(\omega^T (x_1-x_2))\right] \\
\label{eq4}
&= \mathbb{E}_{\omega\sim\mathbb{P}}\left[  \cos(\omega^T (x_1-x_2)\right]\\
&= \mathbb{E}_{\omega\sim\mathbb{P}} \left[  \cos(\omega^T x_1) \cos(\omega^T x_2) +  \sin(\omega^T x_1) \sin(\omega^Tx_2)\right]\\
&\approx \frac{1}{m} \sum_{k=1}^{m}\left( \cos(\omega_k^T x_1) \cos(\omega_k^T x_2) +  \sin(\omega_k^T x_1) \sin(\omega_k^Tx_2)\right)\\
&= \frac{1}{m} \sum_{k=1}^m \Phi_k(x_1)^T \Phi_k(x_2)\label{approximation}
\end{align}

where $\{\omega_k\}_{k=1}^m\iid \mathbb P$  and we denoted
$$\Phi_k(x_l) =  \begin{pmatrix}
 \cos(\omega_k^T x_l)\\
 \sin(\omega_k^T x_l)
\end{pmatrix}.$$

For a covariate design matrix ${\bf X}\in \mathbb R^{n\times D}$ (with rows corresponding to data vectors $x_1,\ldots,x_n$), and frequency matrix $\Omega\in \mathbb R^{m\times D}$ (with rows coresponding to frequencies $\omega_1,\ldots,\omega_m$), we let $\boldsymbol{\Phi_x} = \left[ \cos({\bf X} \Omega^\top)\; \sin({\bf X} \Omega^\top) \right]$ be a $n\times 2m$ matrix referred to as the feature map of the dataset. The estimated  covariance matrix can be computed as $\widehat{\K} = \frac{1}{m}\boldsymbol{\Phi_x}\boldsymbol{\Phi_x}^T$ which has rank at most $2m$. Substituting $\widehat{\K}$ into \eqref{eq:marginallikelihood} now allows rewriting the determinant and the inverse in terms of the $2m\times 2m$ matrix $\boldsymbol{\Phi_x}^T\boldsymbol{\Phi_x}$, thereby reducing the computational cost of inference from $\mathcal{O}(n^3)$ to $\mathcal{O}(nm^2)$, where $m$ is the number of Monte Carlo samples/frequencies. Typically, $m \ll n$.

In particular, by defining $A=\boldsymbol{\Phi_x}^T\boldsymbol{\Phi_x} + m\frac{\sigma_n^2}{\sigma_f^2} I_{2m}$ where $\sigma_n^2$ is the observation noise variance and $\sigma_f^2=\kappa(0)$ is the kernel scale parameter, and taking $R=\text{chol}(A)$ to be the Cholesky factor of $A$, we first calculate vectors $\alpha_1,\alpha_2$ solving the linear systems of equations $R\alpha_1 =  \boldsymbol{\Phi_x}^T{\bf y}$ and $R^T\alpha_2 = \alpha_1$. The log marginal likelihood can then be computed efficiently as:
\begin{equation}
\log p({\bf y}|\theta) = -\frac{1}{2\sigma_n^2}\left(\|{\bf y}\|^2-\|\alpha_1\|^2  \right)-\frac{1}{2}\sum_i \log (R_{ii}^2) + m \log\left(m\frac{\sigma_n^2}{\sigma_f^2}\right) - \frac{n}{2}\log(2\pi\sigma_n^2).
\end{equation}
Additionally, the posterior predictive mean and variance can be estimated as
\begin{eqnarray}
\label{rff:mean_var}
\widehat{\mu_\theta} &=& \frac{\sigma_f^2}{m}\boldsymbol{\Phi_{x^*}}^T \alpha_2\\
\widehat{\sigma_\theta^2} &=& \sigma_n^2 \left(1+\frac{\sigma_f^2}{m} \|\alpha_2\|^2\right).
\end{eqnarray}

There are two important disadvantages of standard random Fourier features as proposed by \cite{Rahimi2008RandomMachines}: firstly, only stationary (shift invariant) kernels can be approximated, and secondly we have to select a priori a specific class of kernels and their corresponding spectral distributions (e.g. Table \ref{tab1}). In this paper, we address both of these limitations, with a goal to construct methods to learn a nonstationary kernel from the data, while preserving the computational efficiency of random Fourier features.

While we can think about the quantities in \eqref{rff:mean_var} as giving approximations to the full GP inference with a given kernel $k$, they are in fact performing exact GP calculations for another kernel $\hat k$ defined using the explicit feature map $\boldsymbol{\Phi_x}$ defined through frequencies sampled from the spectral measure of $k$. We can thus think about these feature maps as parametrizing a family of kernels in their own right and treat frequencies $\omega_1,\ldots,\omega_m$ as kernel parameters to be optimized, i.e. learned from the data by maximizing the log marginal likelihood. It should be noted that dropping the imaginary part of our kernel symmetrizes the spectral measure allowing us to use any $\mathbb{P}(d\omega)$ -- regardless of its symmetry properties, we will still have a real-valued kernel. In particular, one can use an empirical spectral measure defined by any finite set of frequencies.

\subsection{Nonstationary random Fourier features}
Contrary to stationary kernels, which only depend on the lag vector i.e. $\delta=x_i-x_j$, nonstationary kernels depend on the inputs themselves. A simple example of a nonstationary kernel would be the polynomial kernel defined as:
\begin{equation}
k(x_1,x_2) = (x_1^\top x_2+1)^r.
\end{equation}
To extend the stationary random feature mapping to nonstationary kernels, following \cite{Samo2015GeneralizedKernels2,genton2001classes,Yaglom1987CorrelationFunctions}, we will need to use a more general spectral characterisation of positive definite functions which encompasses stationary \emph{and} nonstationary kernels.

\begin{theorem}(Yaglom, 1987 \cite{Yaglom1987CorrelationFunctions,genton2001classes})
	A nonstationary kernel $k(x_1,x_2)$ is positive definite in $\mathbb{R}^d$ if and only if it has the form:
	\begin{align}
	\label{nonstat}
	k(x_1,x_2) = \int_{\mathbb{R}^D \times\mathbb{R}^D} e^{ i (w_1^Tx_1-w_2^Tx_2)} \mu(dw_1, dw_2)
	\end{align}
	
	where $\mu(dw_1, dw_2)$ is the Lebesgue-Stieltjes measure associated to some positive semi-definite function $f(w_1,w_2)$ with bounded variation.
\end{theorem}

From the above theorem, a nonstationary kernel can be characterized by a spectral measure $\mu(d\omega_1, d\omega_2)$ on the product space $\mathbb{R}^D\times\mathbb{R}^D$. Again, without loss of generality we can assume that $\mu$ is a probability measure. If $\mu$ is concentrated along the diagonal, $\omega_1=\omega_2$, we recover the spectral representation of stationary kernels in the previous section. However, exploiting this more general characterisation, we can construct feature mappings for nonstationary kernels. 

Just like in the stationary case, we can approximate (\ref{nonstat}) using Monte Carlo integration. In order to ensure a valid positive semi-definite spectral density we first have to symmetrize $f(\omega_1, \omega_2)$ by ensuring $f(\omega_1, \omega_2) = f(\omega_2, \omega_1)$ and including the diagonal components $f(\omega_1,\omega_1)$ and $f(\omega_2,\omega_2)$ \cite{Remes2017Non-StationaryKernels}. We can take a general form of density $g$ on the product space and ``symmetrize'': 
$$f(\omega_1, \omega_2) = \frac{1}{4} \left(g(\omega_1, \omega_2) + g(\omega_2, \omega_1) + g(\omega_1, \omega_1)+ g(\omega_2, \omega_2) \right).$$

Once again using Monte Carlo integration we get:
\begin{align*}
 k(x_1,x_2)
 &=\frac{1}{4}\int_{\mathbb{R}^D\times\mathbb{R}^D} \left( e ^{ i (\omega_1^Tx_1-\omega_2^Tx_2)} + e ^{ i (\omega_2^Tx_1-\omega_1^Tx_2)}+e ^{ i (\omega_1^Tx_1-\omega_1^Tx_2)}+e ^{ i (\omega_2^Tx_1-\omega_2^Tx_2)} \right) \mu( d\omega_1d\omega_2)\\ 
 &= \frac{1}{4} \mathbb{E}_\mu \left( e ^{ i (\omega_1^Tx_1-\omega_2^Tx_2)} + e ^{ i (\omega_2^Tx_1-\omega_1^Tx_2)}+e ^{ i (\omega_1^Tx_1-\omega_1^Tx_2)}+e ^{ i (\omega_2^Tx_1-\omega_2^Tx_2)} \right)\\
 &\approx \frac{1}{4m} \sum_{k=1}^{m} \left( e ^{ i (x_1^T\omega_k^1-x_2^T\omega^2_k)} + e ^{ i (x_1^T\omega_k^2-x_2^T\omega^1_k)} + e ^{ i (x_1^T\omega_k^1-x_2^T\omega^1_k)}+ e ^{ i (x_1^T\omega_k^2-x_2^T\omega^2_k)} \right)\\
 &= \frac{1}{4m} \sum_{k=1}^{m} \Big\{ \cos(x_1^T\omega_k^1) \cos(x_2^T\omega_k^1) +  \cos(x_1^T\omega_k^1) \cos(x_2^T\omega_k^2) \\ &\qquad\qquad+  \cos(x_1^T\omega_k^2) \cos( x_2^T\omega_k^1) +  \cos(x_1^T\omega_k^2) \cos(x_2^T\omega_k^2)
 \\ &\qquad\qquad+  \sin(x_1^T\omega_k^1) \sin(x_2^T\omega_k^1) +  \sin(x_1^T\omega_k^1) \sin(x_2^T\omega_k^2)
 \\ &\qquad\qquad+  \sin(x_1^T\omega_k^2) \sin(x_2^T\omega_k^1) +  \sin(x_1^T\omega_k^2) \sin(x_2^T\omega_k^2)\Big\} \quad \textit{(taking the real part)}\\
 &= \frac{1}{4m}\sum_{k=1}^{m} \Phi_k(x_1)^T\Phi_k(x_2)
\end{align*}
where $\{(\omega_k^1,\omega_k^2)\}_{k=1}^m\iid \mu$ and
$$\Phi_k(x_l) =  \begin{pmatrix}
 \cos(x_l^T\omega_k^1)+ \cos(x_l^T\omega_k^2)\\
 \sin(x_l^T\omega_k^1)+ \sin(x_l^T\omega_k^2)
\end{pmatrix}.$$

Hence, by denoting $\Omega^l \in \mathbb R^{m\times D}$ (with rows coresponding to frequencies $\omega^l_1,\ldots,\omega^l_m$) for $l=1,2$ as before, we obtain the corresponding feature map for the approximated kernel as an $n\times 2m$ matrix
\begin{align}\label{feat}
x \rightarrow  \boldsymbol{\Phi_x}= \left[
 \cos({\bf X}(\Omega^1)^T)+ \cos({\bf X}(\Omega^2)^T)\;|\;
 \sin({\bf X}(\Omega^1)^T)+ \sin({\bf X}(\Omega^2)^T) 
\right]
\end{align}
and can be condensed to an identical form as in the stationary case.
\begin{align}\label{khat}
\widehat{\K} = \frac{1}{4m}\boldsymbol{\Phi_x}\boldsymbol{\Phi_x}^T.
\end{align}

The non-stationarity in equation \eqref{khat} arises from the product of differing locations $x_1$ and $x_2$ by different frequencies $\omega_k^1, \omega_k^2$, hence making the kernel dependent on the values of $x_1$ and $x_2$ and not only the lag vector. If the frequencies were exactly the same we just revert back to the stationary case. The complete construction of random Fourier feature approximation is summarized in the algorithm below. 

\begin{algorithm}
	\caption{Random Fourier features for nonstationary kernels }
	\begin{algorithmic}[ht]
		\State \textbf{Input:} spectral measure $\mu$, dataset $\bf X$, number of frequencies $m$ 
		\State \textbf{Output:} Approximation to $\K$
		\State \textbf{Start Algorithm:}
		\State Sample $m$ pairs of frequencies $\{(\omega_k^1,\omega_k^2)\}_{k=1}^m\iid \mu$ giving $\Omega^1$ and $\Omega^2$
		\State Compute $\boldsymbol{\Phi_x}= \left[
 \cos({\bf X}(\Omega^1)^T)+ \cos({\bf X}(\Omega^2)^T)\;|\;
 \sin({\bf X}(\Omega^1)^T)+ \sin({\bf X}(\Omega^2)^T) 
\right]\in\mathbb R^{n\times 2m}
$
		\State $\widehat{\K} = \frac{1}{4m}\boldsymbol{\Phi_x}\boldsymbol{\Phi_x}^T$ 
		\State \textbf{End Algorithm}
	\end{algorithmic}
\end{algorithm}
\pagebreak

However, just like in the stationary case,  we can think about nonstationary Fourier feature maps as parametrizing a family of kernels and treat frequencies $\{(\omega_k^1,\omega_k^2)\}_{k=1}^m$ as kernel parameters to be learned by maximizing the log marginal likelihood, which is an approach we pursue in this work. Again, symmetrization due to dropping imaginary parts implies that any empirical spectral measure is valid (there are no constraints on the frequencies).

\subsection{On the choice of spectral measure in non-stationary case}

Using the characterisation in equation \eqref{khat} one only requires the specification of the (Lebesgue-Stieltjes measurable) distribution $f(\omega_1,\omega_2)$ in order to construct a nonstationary kernel. This very general formulation allows us to create the full spectrum encompassing both simple and highly complex kernels.

In the simplest case, $f(\omega_1,\omega_2)=f_1(\omega_1)f_2(\omega_2)$, i.e. it can be a product of popular spectral densities listed in Table \ref{tab1}. Furthermore, one could consider cases where these individual spectral densities factorize further across dimensions. This corresponds to a notion of \emph{separability}. In spatio-temporal data, separability can be very useful as it enables interpretation of the relationship between the covariates as well as computationally efficient estimations and inferences \cite{Finkenstadt2007StatisticalSystems}. Practical implementation is straightforward; consider the classic spatio-temporal setting with 3 covariates - longitude, latitude and time. When drawing random samples of $\omega_l = (\omega_l^1, \omega_l^2, \omega_l^3)$ where $l\in\{1,2\}$, we could define the $\omega_l^i$ to come from different distributions, allowing us to individually model each input dimension. If the distribution on frequencies are independent across dimensions then we see that if $\omega_1 = (\omega_1^1, \omega_1^2, \omega_1^3)$ and $\omega_2 = (\omega_2^1, \omega_2^2, \omega_2^3)$:
\begin{align}\label{sepcase}
	k(x_1,x_2) &= \int e^{i\omega_1^T x_1-i\omega_2^T x_2}f(\omega_1,\omega_2)d\omega_1d\omega_2\\
    &= \int e^{i(\omega^1_1 x_1^1 + \omega_1^2 x_1^2 + \omega^3_1 x_1^3-\omega^1_2 x_2^1 - \omega_2^2 x_2^2 - \omega^3_2 x_2^3)}\prod_{p=1}^3f(\omega_1^p, \omega_2^p) d\omega_1^pd\omega_2^p \\
	&= k_1(x_1^1,x_2^1)k_2(x_1^2,x_2^2)k_3(x_1^3,x_2^3).
\end{align}

A practical example for spatio-temporal modelling of a nonstationary separable kernel would be generating a four dimensional ($\omega_1^1,\omega_2^1,\omega_1^2,\omega_2^2$) , sample from independent Gaussian distributions (whose spectral density corresponds to a squared exponential kernel) representing nonstationary spatial coordinates, and a two dimensional ($\omega_1^3,\omega_2^3$) from a Student-t distribution with $0.5$ degrees of freedom (whose spectral density corresponds to a Mat\'ern $1/2$ kernel or exponential kernel) representing temporal coordinates.

To move from separable to non-separable, nonstationary kernels one only needs to introduce some dependence structure within  $\omega^1$ or $\omega^2$ i.e. across feature dimensions, such as for example using the multivariate normal distribution in $\mathbb{R}^{D}$, in order to prevent the factorization in equation \eqref{sepcase}. The correlation structure in these multivariate distributions are what creates the non-separability.

To create non-separable kernels with different spectral densities along each feature dimension copulas can be used. An example in a spatial (latitude, longitude feature dimensions) setting using the Gaussian copula, would involve generating samples for $\omega^1$ or $\omega^2 \in \mathbb{R}^2$ (or both) from a multivariate normal distribution $\{\omega_k^1\}_{k=1}^m\iid \mathcal{N}(0,\Sigma)$, pass these through the Gaussian cumulative distribution function, and then used in the quantile function of another distribution ($\Lambda$) i.e. $\mathcal{C}_\Lambda(\omega^1) = CDF_\Lambda(CDF_{\mathcal{N}}^{-1}(\omega^1))$. This transformation can also be done using different $\Lambda$s along different feature dimensions. Alternative copulas can be readily used, including the popular Archimedian Copulas: Clayton, Frank and Gumbel\cite{Genest1986TheMarginals}. Additionally, mixtures of multivariate normals can be used \cite{Remes2017Non-StationaryKernels,Yang2015AKernels} to create arbitrarily complex non-separable and nonstationary kernels. Given sufficient components any probability density function can be approximated to the desired accuracy.

In this paper, we focus on the most general case where the frequencies $\{(\omega_k^1,\omega_k^2)\}_{k=1}^m$ are treated as kernel parameters and are learnt directly from the data by optimising the marginal likelihood, i.e. they are not associated to any specific family of joint distributions. This approach allows us to directly learn nonstationary kernels of arbitrary complexity as $m$ increases. However, a major problem with such a heavily overparametrized kernel is the possibility of overfitting. Stationary examples of learning frequencies directly from the data \cite{Lazaro-Gredilla2010SparseRegression,Gal2015ImprovingInputs,Tan2013VariationalRegression} have been known to overfit despite the regularisation due to working with marginal likelihood. This problem is further exacerbated in high-dimensional settings, such as those in spatio-temporal mapping with covariates. In this paper, we include an additional regularisation inspired by dropout \cite{Srivastava2014} which prevents the co-adaptation of the learnt frequencies $\omega_1,\omega_2$.

\subsection{Gaussian dropout regularisation}

Dropout \cite{Srivastava2014} is a regularisation technique introduced to mitigate overfitting in deep neural networks. In its simplest form, dropout involves setting features/matrix entries to zero with probability $q = 1-p$, i.e. according to a $Bernoulli(p)$ for each feature. The main motivation behind the algorithm is to prevent co-adaptation by forcing features to be robust and rely on population behaviour. This prevents individual features from overfitting to idiosyncrasies of the data.

\begin{figure}[htbp] \centering  \includegraphics[width=0.8\textwidth]{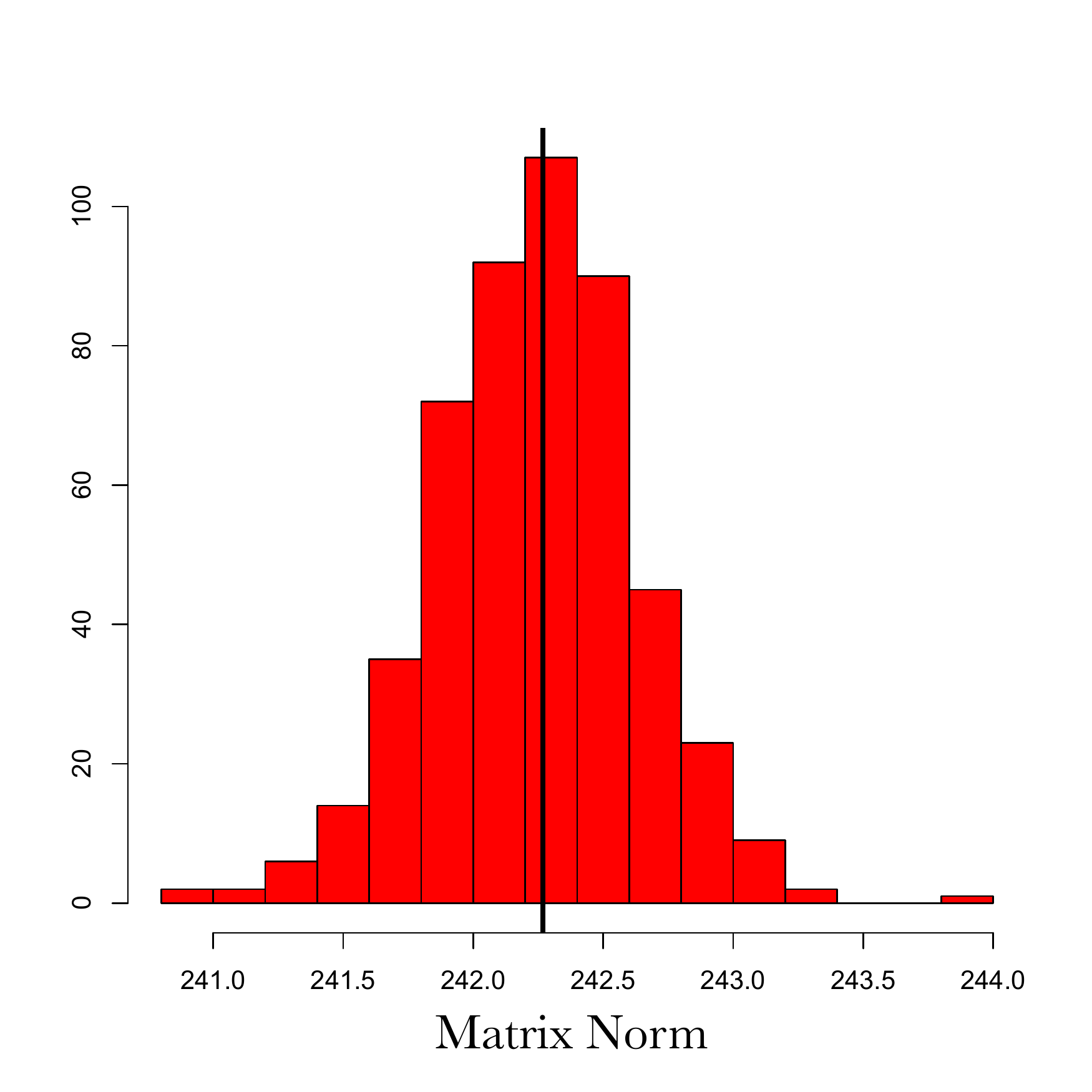} \caption{\small Histogram of the Euclidean norm of a covariance matrix $\Phi\Phi^T$ with Gaussian dropout of $\sigma_p = 0.05$. The black line is the norm of $\Phi\Phi^T$ \emph{without} noise} \label{fig:dropfig} \end{figure}

Using standard dropout, where zeros are introduced into the frequencies $\{(\omega_k^1,\omega_k^2)\}_{k=1}^m$ can be problematic due to the trigonometric transformations in the projected features. An alternative to dropout that has been shown to be just as effective if not better is Gaussian dropout\cite{Srivastava2014,Baldi2013UnderstandingDropout}. Regularisation via Gaussian dropout involves augmenting our sample distribution as $\{(\omega_k^1,\omega_k^2)_\eta\}_{k=1}^m= \mathcal{N}(1,\sigma_{p}^2)\odot\{(\omega_k^1,\omega_k^2)\}_{k=1}^m$. The addition of noise through $\mathcal{N}(1,\sigma_{p}^2)$ ensures unbiased estimates of the covariance matrix i.e $\mathbb{E}[\{(\omega_k^1,\omega_k^2)_\eta\}_{k=1}^m]=\mathbb{E}[\{(\omega_k^1,\omega_k^2)\}_{k=1}^m]$ (see figure \ref{fig:dropfig}). As with dropout, this approach prevented our population Monte Carlo sample from co-adapting, and ensured that the learnt frequencies are robust and not overfitting noise in the data. An additional benefit of this procedure over improving generalisation error and preventing overfitting was to speed up the convergence of gradient descent optimisers through escaping saddle points more effectively \cite{Lee2017First-orderPoints}. The noise parameter $\sigma_{p}$ defines the degree of regularisation and is a hyperparameter that needs to be tuned. However, we found in practice when coupled with an early stopping procedure, learning the frequencies is robust to sensible choices of $\sigma_{p}$. 

\section{Results}

\subsection{Google daily high stock price}

To demonstrate the use of the developed method and the utility of nonstationary modelling, we consider time series data of the daily high stock price of Google spanning $3295$ days from 19th August 2004 to 20th September 2017. We set $x\in\{1,\ldots,3295\}$ and $y=\log(Stock_{high})$. For the stationary case we use vanilla random Fourier features\cite{Rahimi2008RandomMachines,Lazaro-Gredilla2010SparseRegression} with the squared exponential kernel (Gaussian spectral density) and $m=600$ \emph{fixed} frequencies. For the nonstationary case we use $m=300$ frequencies  for each $\omega_1$ and for $\omega_2$. We performed a sensitivity analysis to check that no improvements in either the log marginal likelihood or testing error resulted from using more features. Cross-validation was performed using a $70-30$ training-testing split averaged over 20 independent runs and testing performance evaluated via the mean squared error and correlation. Optimisation of the log marginal likelihood was performed using ADAM\cite{Kingma2014Adam:Optimization} gradient descent in TensorFlow \cite{Abadi2016TensorFlow:Systems} with early stopping and patience \cite{Prechelt1998EarlyWhen}.

\begin{figure*}
        \centering
        \begin{subfigure}[b]{0.475\textwidth}
            \centering
            \includegraphics[width=\textwidth]{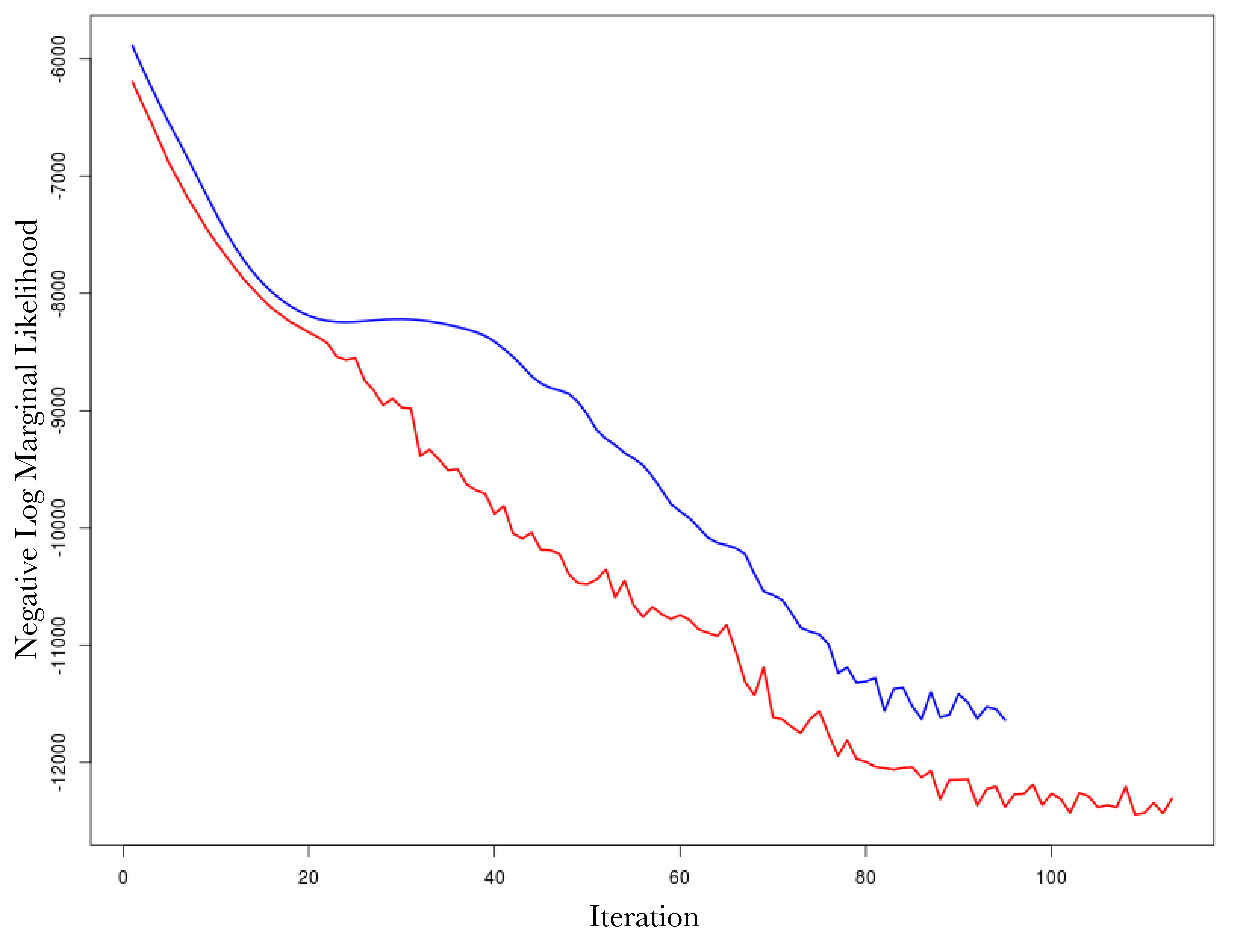}   
            \label{fig:1a}
        \end{subfigure}
        \hfill
        \begin{subfigure}[b]{0.475\textwidth}  
            \centering 
            \includegraphics[width=\textwidth]{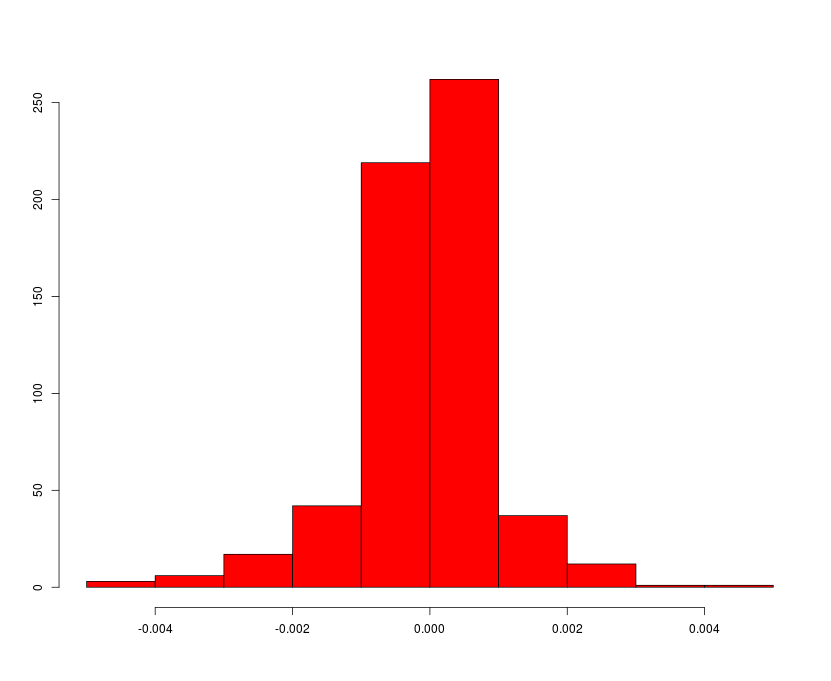}   
            \label{fig:1b}
        \end{subfigure}
        \vskip\baselineskip
        \begin{subfigure}[b]{0.475\textwidth}   
            \centering 
            \includegraphics[width=\textwidth]{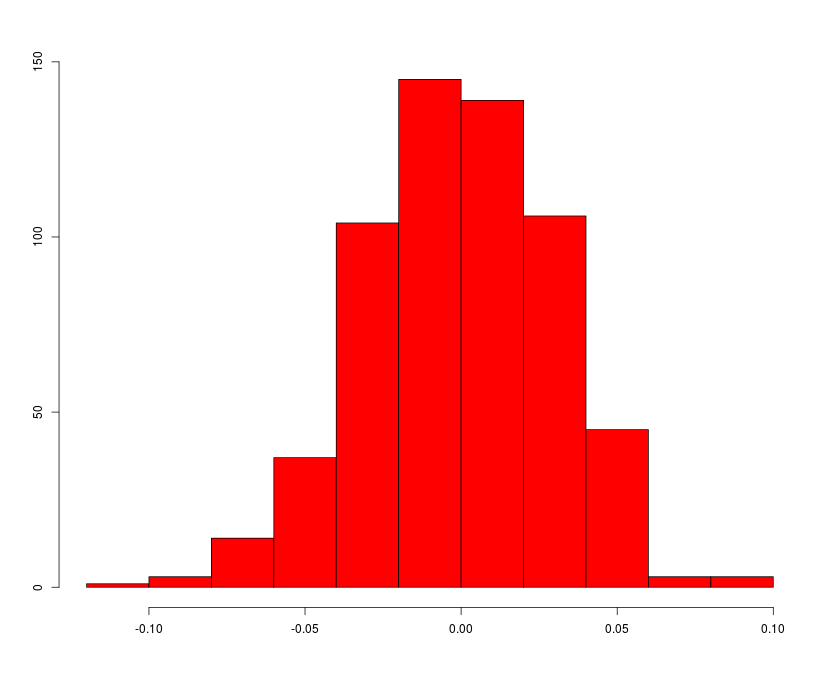}
            \label{fig:1c}
        \end{subfigure}
        \quad
        \begin{subfigure}[b]{0.475\textwidth}   
            \centering 
            \includegraphics[width=\textwidth]{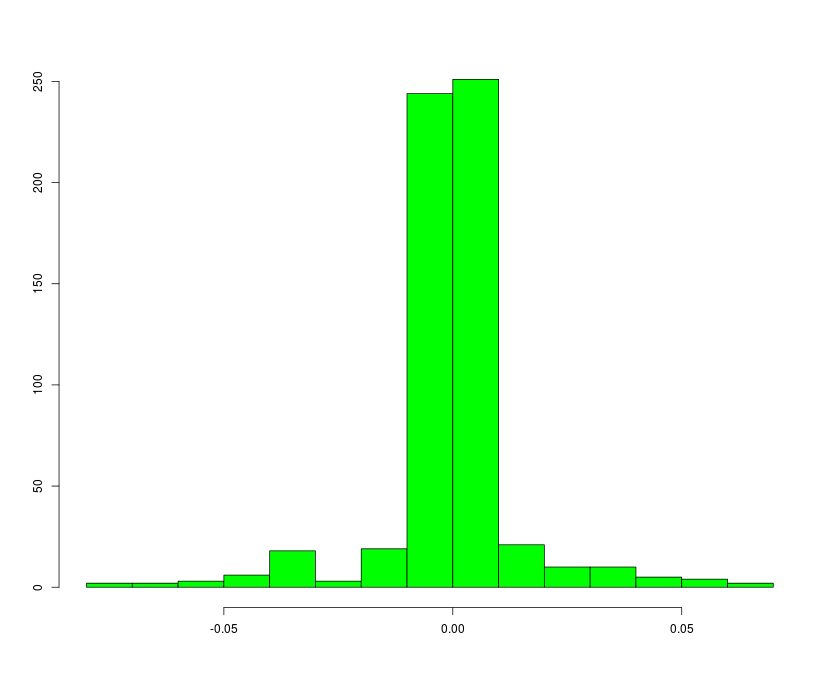} 
            \label{fig:1d}
        \end{subfigure}
   		  \caption[]%
            {{\small(top left) Log Marginal likelihood (Y-axis), optimisation gradient 	update  count (X-axis), vanilla random  features (blue), proposed approach (red); (top right) Histogram of learnt $\omega_1$ for our nonstationary approach; (bottom left) Histogram of learnt $\omega_2$ for our nonstationary approach; (bottom right) Histogram of $\omega_2$ for vanilla random Fourier features}}           
        \label{fig:fig1}
    \end{figure*}

Figure \ref{fig:fig1} (top left) shows the comparison in the optimisation paths of the negative log marginal likelihood between the two methods. It is clear that the nonstationary approach reaches a lower minima than the vanilla random Fourier features approach. This is also mirrored in the testing accuracy over the 20 independent runs where our approach achieves a mean squared error and correlation of $3.29\times 10^{-5}$ and $0.999$, while the vanilla Fourier features approach  achieves a mean squared error and correlation of $5.69\times 10^{-5}$ and $0.987$. Of note is the impact of our Gaussian dropout regularisation, which, through the injection of noise, appears to converge faster and avoid plateaus. This is entirely in keeping with previous experiences using dropout variants\cite{Baldi2013UnderstandingDropout} and highlights an added benefit over using only ridge (weight decay) regularisation.

\begin{figure*}
        \centering
        \begin{subfigure}[b]{1.1\textwidth}
            \centering
            \includegraphics[width=\textwidth]{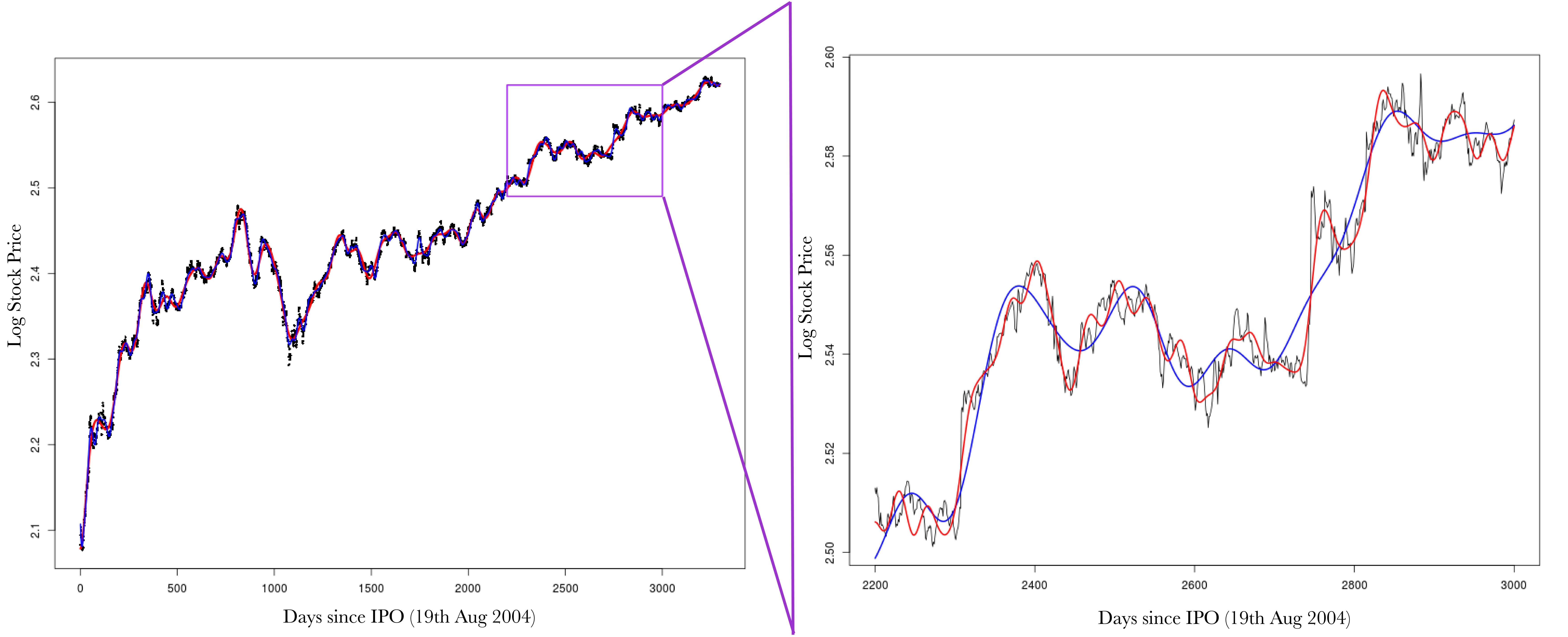}   
            \label{fig:2a}
        \end{subfigure}
        \vskip\baselineskip
        \begin{subfigure}[b]{0.475\textwidth}  
            \centering 
            \includegraphics[width=\textwidth]{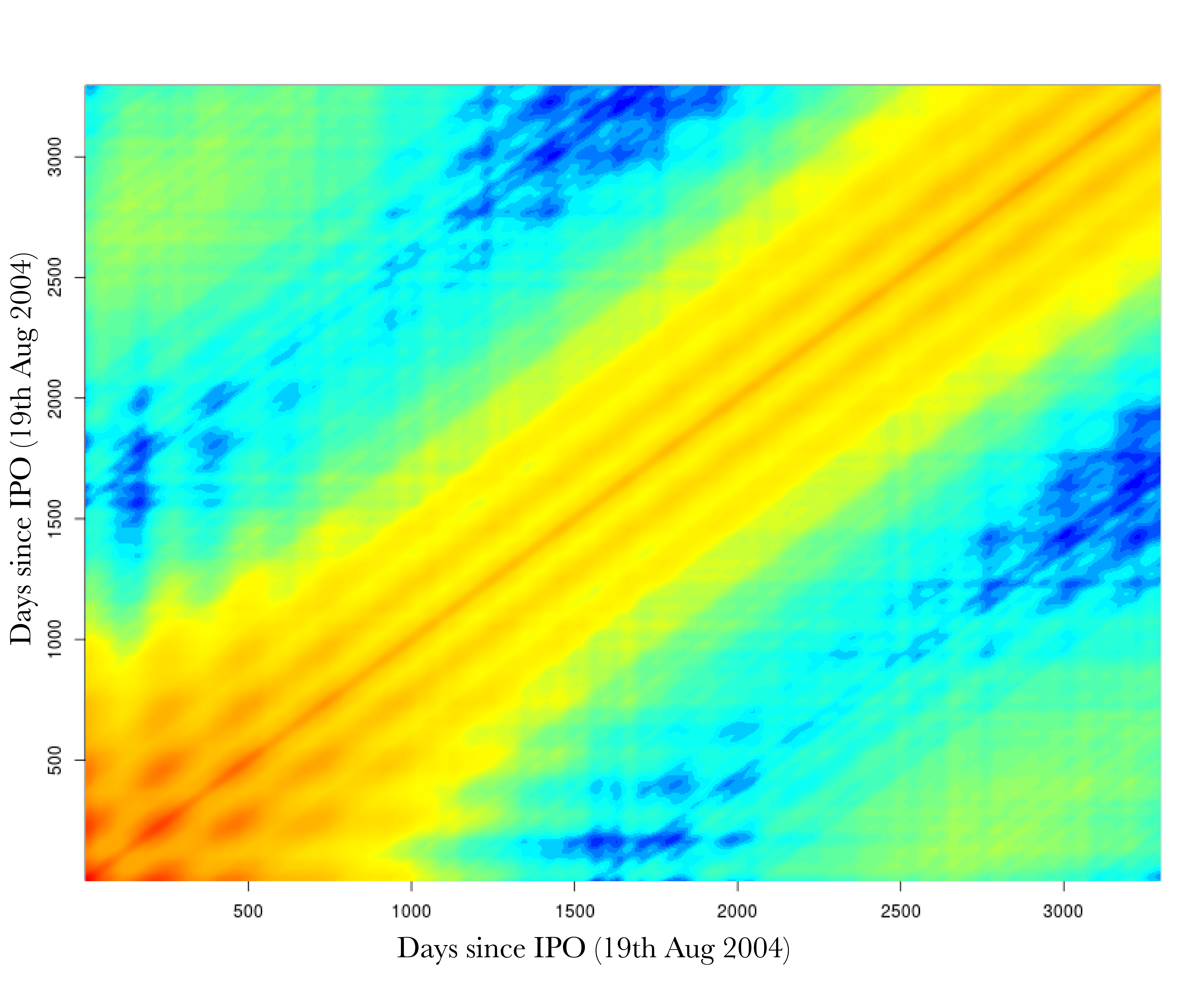}   
            \label{fig:2b}
        \end{subfigure}
        \begin{subfigure}[b]{0.475\textwidth}   
            \centering 
            \includegraphics[width=\textwidth]{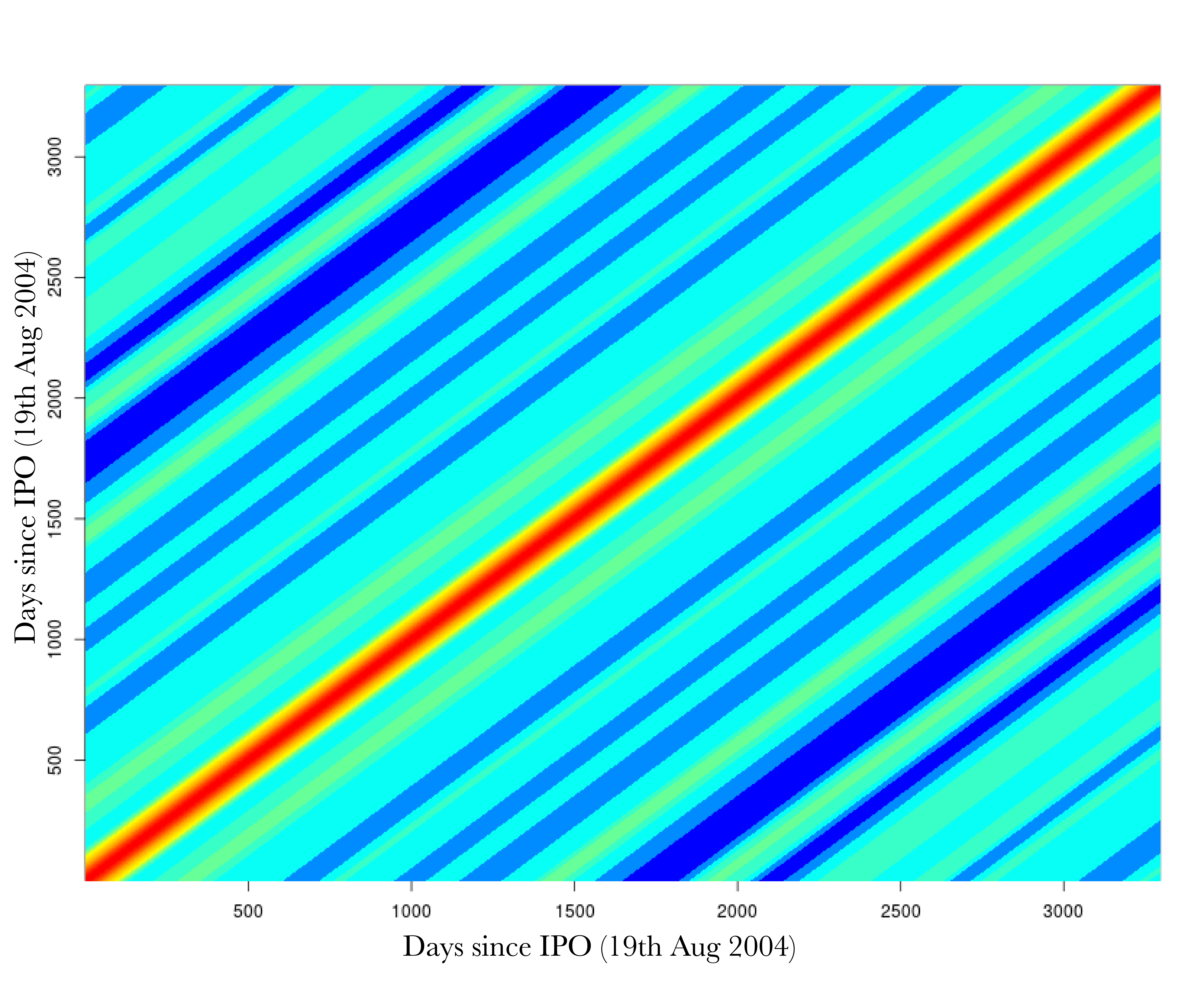}
            \label{fig:2c}
        \end{subfigure}
   		  \caption[]%
            {{\small(top) Log daily-high Google stock price (Y-axis), days since 19th August 2004  (X-axis). Vanilla random  features (blue) our proposed approach(red), actual data (black), with a zoomed in transect (purple box); (bottom left) Image of covariance matrix for our nonstationary method; (bottom right) Image of covariance matrix for vanilla random Fourier features}}           
        \label{fig:fig2}
    \end{figure*}

Figure \ref{fig:fig2} (top) shows the overall fits compared to the raw data. Both methods appear to fit the data very well, as reflected in the testing statistics, but when examining a zoomed-in transect it is clear that the learnt nonstationary features fit the data better than the vanilla random features by allowing variable degree of smoothness in time. The combination of nonstationarity and kernel flexibility allowed us to learn a much better characterisation of the data patterns without overfitting. The covariance matrix comparisons (Figure \ref{fig:fig2} bottom) further highlight this point where the learnt nonstationary covariance matrix shares some similarities with the vanilla random features covariance matrix, such as the concentration on the diagonal, but exhibits a much greater degree of texture.  The histograms in Figure \ref{fig:fig2} provide another perspective on the covariance structure, where the vanilla features are by design Gaussian distributed, but learnt nonstationary frequencies are far from Gaussian (Kolmogorov-Smirnov test p-value$<10^{-16}$) exhibiting skewness and heavy tails. Additionally, the differences between the learnt frequencies $\omega_1$ and $\omega_2$ show that, not only is the learnt kernel far from Gaussian, but that it is indeed also nonstationary. This simple example also highlights the potential defficiencies of choosing kernels/frequencies $\emph{a priori}$.

\subsection{Spatial temperature anomaly for East Africa in 2016}

\begin{figure}[htbp] \centering  \includegraphics[width=1.1\textwidth]{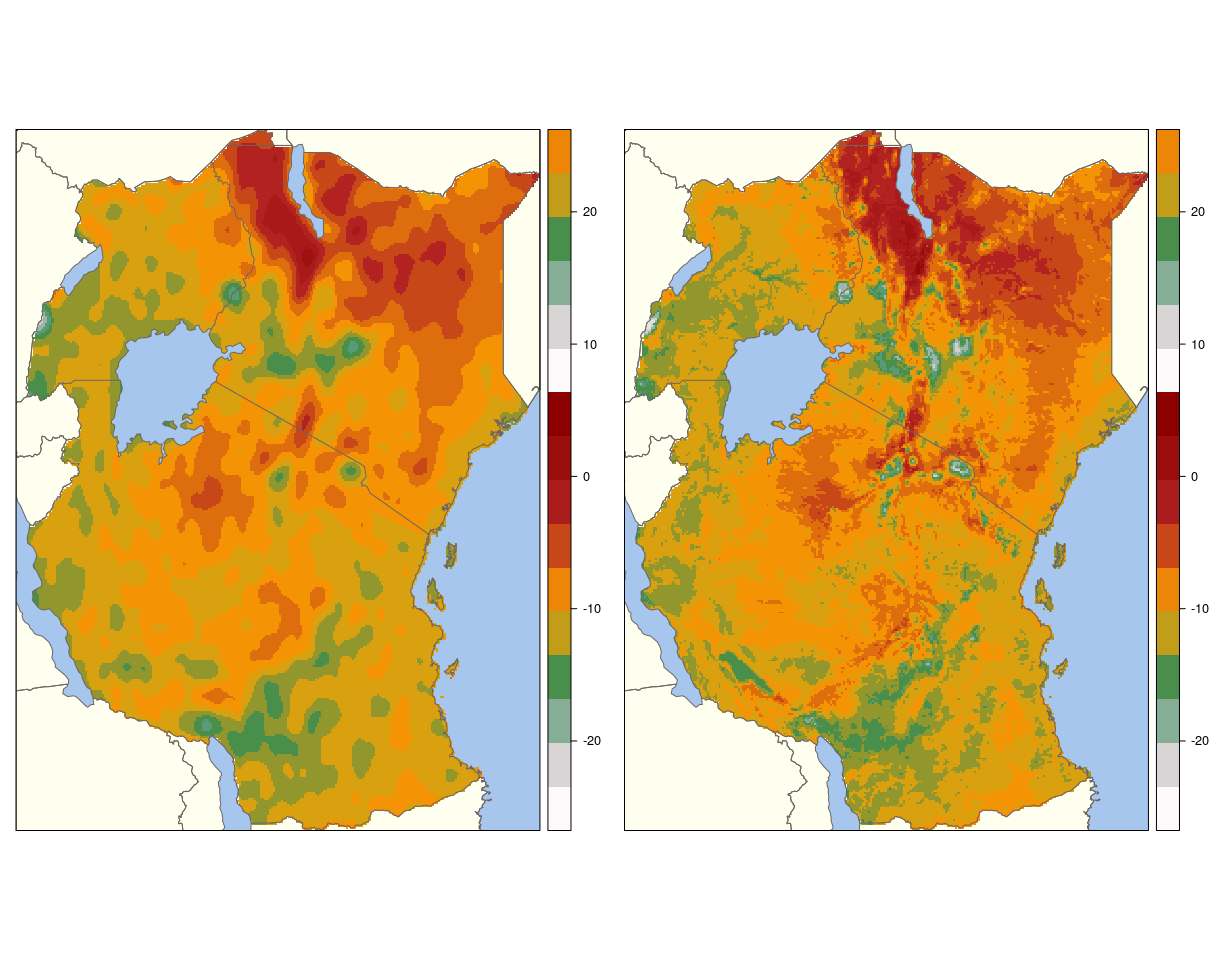} \caption{\small Predicted (left) verses actual (right) temperature anomaly maps} \label{fig:fig4} \end{figure}

\begin{figure}[hbtp] \centering  \includegraphics[width=1.1\textwidth]{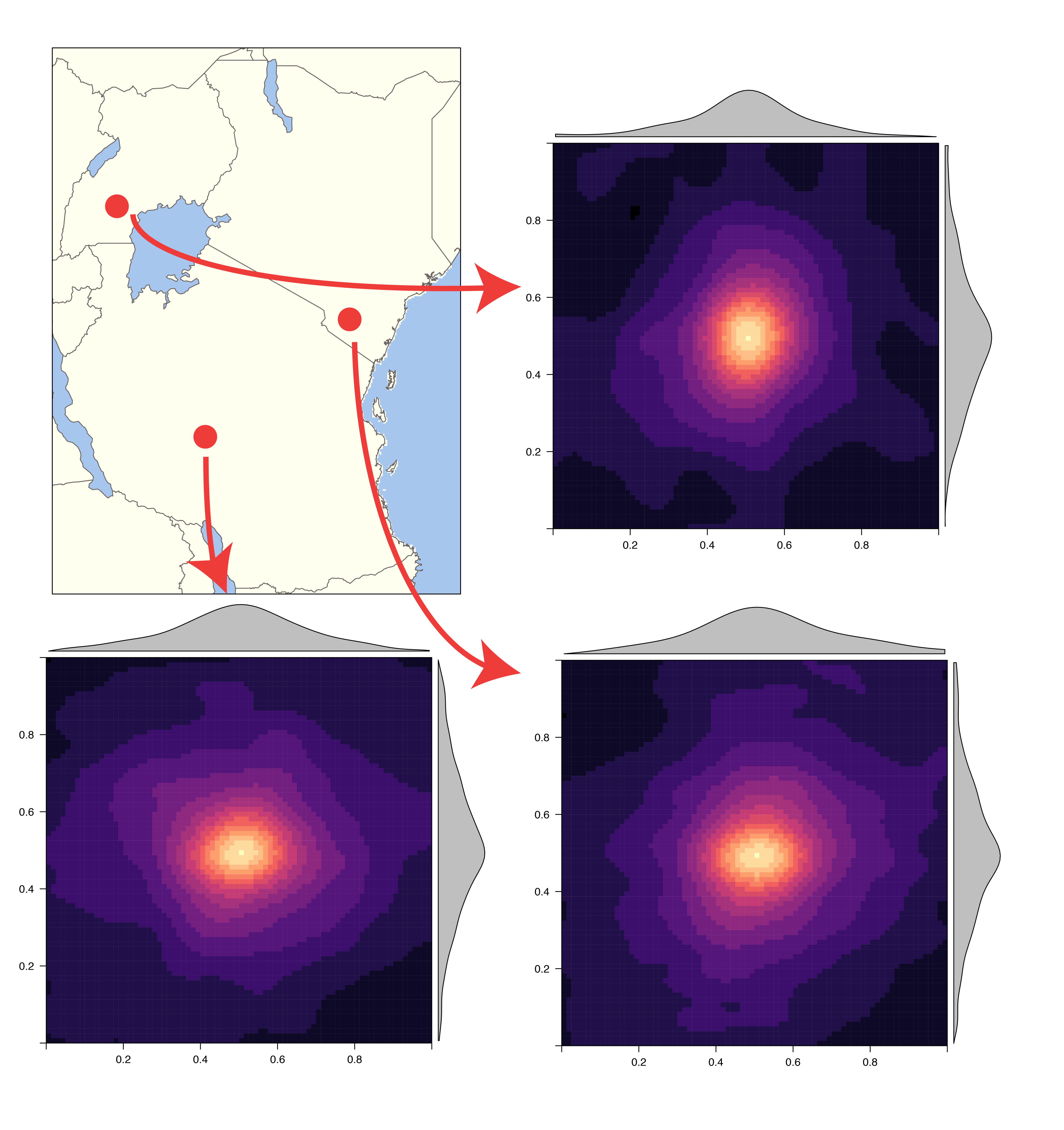} \caption{\small Covariance matrix images for 3 random points showing different covariance structures due to nonstationarity} \label{fig:fig5} \end{figure}

\begin{figure}[hbtp] \centering  \includegraphics[width=1.1\textwidth]{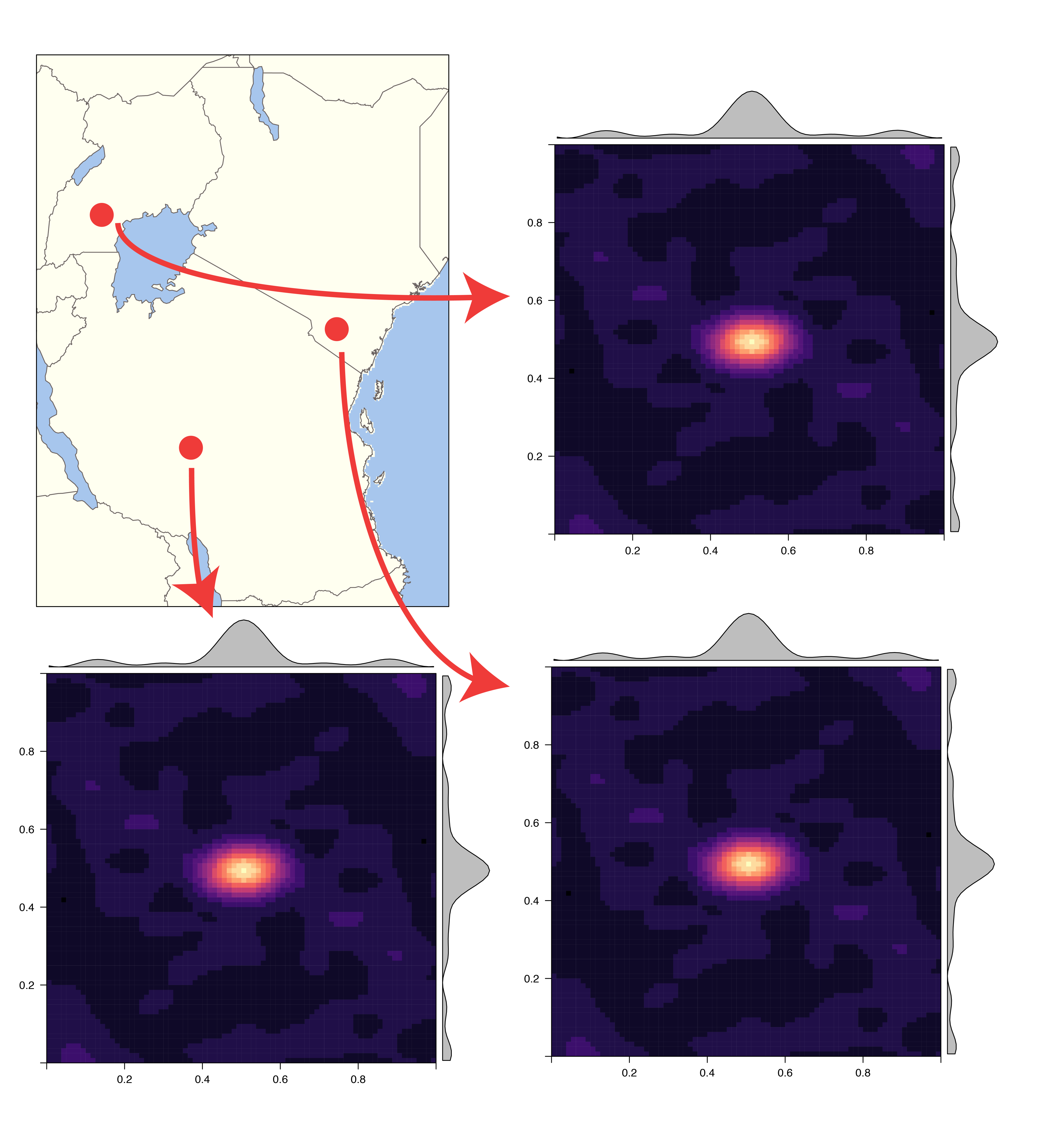} \caption{\small Covariance matrix images for 3 random points showing identical covariance structures for all locations due to stationarity} \label{fig:fig6} \end{figure}

We next consider MOD11A2 Land Surface Temperature (LST) 8-day composite data \cite{Wan2002ValidationData}, gap filled for cloud cover images \cite{Weiss2014AnTime-series} and averaged to a synoptic yearly mean for 2016. To replicate common situations for spatial mapping, such as interpolation from sparse remote sensed sites or cluster house hold survey locations \cite{Bhatt2017} we randomly sample $4000$ LST locations (only $\sim 4\%$ of the total) from the East Africa region (see Figures \ref{fig:fig4} and \ref{fig:fig5}). We set $x\in\mathbb{R}^2=\{Latitude,Longitude\}$ i.e. using only the spatial coordinates as covariates, and use the LST temperature anomaly as the response. We apply our nonstationary approach, learning $m=600$ frequencies for $\omega_1$ and $\omega_2$ each. Cross validation was evaluated over all pixels excluding the training set ($\sim 83,000$) and averaged over 10 independent runs with testing performance evaluated via the mean squared error and correlation. Our final performance estimates were $4.23$ and $0.91$ for the mean squared error and correlation respectively. Figure \ref{fig:fig4} shows our predicted surface (left) compared to the actual data (right). Our model shows strong correspondence to the underlying data and highlights the suitability of using our approach in settings where no relevant covariates exist outside of the spatial coordinates. Figure \ref{fig:fig5} shows 3 randomly sampled points and the covariance patterns around those points. For comparison Figure \ref{fig:fig6} shows the equivalent plot when using a stationary squared exponential kernel. In stark contrast to the stationary covariance function, which has an identical structure for all three points, the nonstationary kernel shows considerable heterogeneity in both patterns and shapes. Interestingly the learnt lengthscale/bandwidth seems to be much smaller in the stationary case than the nonstationary case, we hypothesise that this is due to the inability of the stationary kernel to learn the rich covariance structure needed to accurately model temperature anomaly. Intuitively, nonstationarity allows locally dependent covariance structures which conform to the properties of a particular location and imply (on average) larger similarity of nearby outputs and better generalisation ability. In contrast, stationary kernels are trying to fit one covariance structure to all locations and as a result end up with a much shorter lengthscale as it needs to apply to all directions from all locations. Our results are in concordance with other studies showing that temperature anomaly data is nonstationary\cite{Remes2017Non-StationaryKernels,Samo2015GeneralizedKernels2,Wu2007OnSeries.}. This interpolation problem can readily be expanded into multiple dimensions including time and other covariates.

\section{Discussion}

We have shown that nonstationary kernels of arbitrary complexity are as easy to model as stationary ones, and can be learnt with sufficient efficiency to be applicable to datasets of all sizes. The qualitative superiority of predictions when using nonstationary kernels has previously been noted\cite{Paciorek2006SpatialFunctions}. In many applications, such as in epidemiology where data can be noisy, generalisation accuracy is not the only measure of model performance, and there is a need for models that conform to known biological constraints and external field data. The flexibility of nonstationary kernels allows for more plausible realities to be modelled without the assumption of stationarity limiting the expressiveness of the predictions.  It has also been noted that while nonstationary GPs give more sensible results than stationary GPs, they often show little generalisation improvement \cite{Paciorek2006SpatialFunctions}. For the examples in this work we show clear improvements in generalisation accuracy when using nonstationary kernels. We conjecture that the differences in generalisation performance are likely due to the same reasons limiting neural network performance a decade ago\cite{LeCun2015DeepLearning} - namely, a combination of small, poor quality data and a lack of generality in the underlying specification. Given more generalised specifications, such as those introduced in this paper, coupled with the current trend of increasing quantities of high quality data\cite{2014TrendsAnalytics} we believe nonstationary approaches will be more and more relevant in spatio-temporal modelling. 

There has long been codes of practice on which kernel to use on which spatial dataset\cite{Diggle2007Model-basedGeostatistics} based on a priori assumptions about the roughness of the underlying process. Using the approach introduced in this paper, \emph{ad hoc} choices of kernel and decisions on stationarity versus nonstationarity may no longer be needed as it may be possible to learn the kernel automatically from the data. For example, our approach can be easily modified to vary the degree of nonstationarity according to patterns found in the data.

In this work we have focused on optimising the marginal likelihood in Gaussian Process regression and added extra regularisation via Gaussian dropout. However, for non-Gaussian observation models the marginal likelihood cannot be obtained in a closed form. In these settings, one may resort to frequentist methods instead and resort to variational\cite{Tan2013VariationalRegression}, approximate\cite{Rue2009ApproximateApproximationsb,Minka2001ExpectationInference} or suitable MCMC\cite{CarpenterStan:Language} approaches in order to provide uncertainty measures. For very large models with non-Gaussian observation models, stochastic gradient descent in mini-batches\cite{Bengio2012PracticalArchitectures} or stochastic gradient Bayesian methods \cite{Chen2014StochasticCarlo} can be used. 

The matrix $\boldsymbol{\Phi}$ that results from the Fourier feature expansion can be thought of as a hidden layer in a single layer neural network. This formulation explicitly connects the random Fourier feature approach and single layer neural networks using learnable (or random) nodes and specific trigonometric activation functions. Our approach can therefore be used in deep learning settings while retaining an explicit link to kernel methods and their interpretability \cite{Lee2017DeepProcesses}.

\pagebreak
\section{Figures}

\clearpage
\nolinenumbers
\section*{Author Contributions} 
Conceived of and designed the research: SB, DS, SF, JFT. Drafted the manuscript: SB, DS, SF, JFT. Conducted the analyses: SB,JFT. Supported the analyses: SB, DS, SF, JFT.All authors discussed the results and contributed to the revision of the final manuscript. 

\subsection*{Funding Statement}
SB is supported by the MRC outbreak centre and the Bill and Melinda Gates Foundation $[\text{OPP}1152978]$.

\section{References}
\bibliography{references}

\begin{thebibliography}{10}

\bibitem{2000APrices}
{A method for spatial–temporal forecasting with an application to real estate
  prices}.
\newblock {\em International Journal of Forecasting}, 16(2):229--246, 4 2000.

\bibitem{Berrocal2007CombiningForecasts}
Veronica~J. Berrocal, Adrian~E. Raftery, Tilmann Gneiting, Veronica~J.
  Berrocal, Adrian~E. Raftery, and Tilmann Gneiting.
\newblock {Combining Spatial Statistical and Ensemble Information in
  Probabilistic Weather Forecasts}.
\newblock {\em Monthly Weather Review}, 135(4):1386--1402, 4 2007.

\bibitem{2002SpatialSensing}
Alfred Stein, Freek Van~der Meer, and Ben Gorte, editors.
\newblock {\em {Spatial Statistics for Remote Sensing}}, volume~1 of {\em
  Remote Sensing and Digital Image Processing}.
\newblock Springer Netherlands, Dordrecht, 2002.

\bibitem{Weiss2014AirPrediction}
D.J. Daniel J~J Weiss, Samir Bhatt, Bonnie Mappin, Thomas P P~T.P. Van~Boeckel,
  David L L~D.L. Smith, S.I. Simon I~I Hay, and Peter W W~P.W. Gething.
\newblock {Air temperature suitability for Plasmodium falciparum malaria
  transmission in Africa 2000-2012: a high-resolution spatiotemporal
  prediction.}
\newblock {\em Malaria journal}, 13(1):171, 1 2014.

\bibitem{Weiss2014AnTime-series}
Daniel J~D.J. Weiss, Peter M~P.M. Atkinson, Samir Bhatt, Bonnie Mappin, Simon
  I~S.I. Hay, and Peter W~P.W. Gething.
\newblock {An effective approach for gap-filling continental scale remotely
  sensed time-series.}
\newblock {\em ISPRS journal of photogrammetry and remote sensing : official
  publication of the International Society for Photogrammetry and Remote
  Sensing (ISPRS)}, 98:106--118, 12 2014.

\bibitem{Bhatt2015The2015}
S.~Bhatt, D.J. J~J Weiss, E.~Cameron, D.~Bisanzio, B.~Mappin, U.~Dalrymple,
  K.E. E~E Battle, C.L. L~L Moyes, A.~Henry, P.A. A~A Eckhoff, E.A. A~A Wenger,
  O.~Bri{\"{e}}t, M.A. A~A Penny, T.A. A~A Smith, A.~Bennett, J.~Yukich, T.P.
  P~P Eisele, J.T. T~T Griffin, C.A. A~A Fergus, M.~Lynch, F.~Lindgren, J.M.
  M~M Cohen, C.L.J. L J L~J Murray, D.L. L~L Smith, S.I. I~I Hay, R.E. E~E
  Cibulskis, and P.W. W~W Gething.
\newblock {The effect of malaria control on Plasmodium falciparum in Africa
  between 2000 and 2015.}
\newblock {\em Nature}, 526(7572):207--211, 9 2015.

\bibitem{Bhatt2013TheDengue}
Samir Bhatt, Peter W~P.W. Gething, Oliver J~O.J. Brady, J.P. Jane~P Messina,
  Andrew W~A.W. Farlow, Catherine L~C.L. Moyes, J.M. John~M Drake, J.S. John~S
  Brownstein, Anne G~A.G. Hoen, Osman Sankoh, M.F. Monica~F Myers, Dylan B~D.B.
  George, Thomas Jaenisch, G.R. William~Wint, C.P. Cameron~P Simmons, Thomas
  W~T.W. Scott, Jeremy~J.J. Farrar, Simon I~S.I. Hay, G~R~William Wint, C.P.
  Cameron~P Simmons, Thomas W~T.W. Scott, Jeremy~J.J. Farrar, and Simon I~S.I.
  Hay.
\newblock {The global distribution and burden of dengue.}
\newblock {\em Nature}, 496(7446):504--7, 4 2013.

\bibitem{hay2013global}
Simon I~S.I. Hay, K.E. Katherine~E Battle, D.M. David~M Pigott, D.L. David~L
  Smith, Catherine L~C.L. Moyes, Samir Bhatt, J.S. John~S Brownstein, Nigel
  Collier, M.F. Monica~F Myers, Dylan B~D.B. George, Peter W~P.W. Gething,
  {others}, and Peter W~P.W. Gething.
\newblock {Global mapping of infectious disease.}
\newblock {\em Philosophical transactions of the Royal Society of London.
  Series B, Biological sciences}, 368(1614):20120250, 3 2013.

\bibitem{Rasmussen2006GaussianLearning}
Carl~Edward Rasmussen and Christopher K~I Williams.
\newblock {\em {Gaussian processes for machine learning}}, volume~1.
\newblock MIT press Cambridge, 2006.

\bibitem{Diggle2007Model-basedGeostatistics}
Peter Diggle and PJ~Ribeiro.
\newblock {\em {Model-based Geostatistics}}.
\newblock Springer, New York, 2007.

\bibitem{Paciorek2006SpatialFunctions}
Christopher~J. Paciorek and Mark~J. Schervish.
\newblock {Spatial modelling using a new class of nonstationary covariance
  functions}.
\newblock {\em Environmetrics}, 17(5):483--506, 8 2006.

\bibitem{genton2001classes}
Marc~G Genton.
\newblock Classes of kernels for machine learning: a statistics perspective.
\newblock {\em Journal of machine learning research}, 2(Dec):299--312, 2001.

\bibitem{Micchelli2006}
Charles~A. Micchelli, Yuesheng Xu, and Haizhang Zhang.
\newblock Universal kernels.
\newblock {\em J. Mach. Learn. Res.}, 7:2651--2667, December 2006.

\bibitem{quinonero2005unifying}
Joaquin Qui{\~{n}}onero-Candela and Carl~Edward Rasmussen.
\newblock {A unifying view of sparse approximate Gaussian process regression}.
\newblock {\em The Journal of Machine Learning Research}, 6(Dec):1939--1959,
  2005.

\bibitem{Snelson2012VariableProcesses}
Edward Snelson and Zoubin Ghahramani.
\newblock {Variable noise and dimensionality reduction for sparse Gaussian
  processes}.
\newblock {\em arXiv preprint arXiv:1206.6873}, 2012.

\bibitem{Lindgren2011AnApproach}
Finn Lindgren, Håvard Rue, and Johan Lindstr{\"{o}}m.
\newblock {An explicit link between Gaussian fields and Gaussian Markov random
  fields: the stochastic partial differential equation approach}.
\newblock {\em Journal of the Royal Statistical Society: Series B (Statistical
  Methodology)}, 73(4):423--498, 2011.

\bibitem{Rahimi2008RandomMachines}
A~Rahimi and B~Recht.
\newblock {Random features for large-scale kernel machines}.
\newblock {\em Advances in neural information processing}, 2008.

\bibitem{Lazaro-Gredilla2010SparseRegression}
Miguel Lázaro-Gredilla, Joaquin~Quiñonero Candela, Carl~Edward Rasmussen, and
  Aníbal~R. Figueiras-Vidal.
\newblock Sparse spectrum gaussian process regression.
\newblock {\em Journal of Machine Learning Research}, 11:1865--1881, 2010.

\bibitem{Samo2015GeneralizedKernels2}
Yves-Laurent~Kom Samo and Stephen Roberts.
\newblock {Generalized Spectral Kernels}.
\newblock 2015.

\bibitem{wilson2013gaussian}
Andrew Wilson and Ryan Adams.
\newblock {Gaussian process kernels for pattern discovery and extrapolation}.
\newblock In {\em Proceedings of the 30th International Conference on Machine
  Learning (ICML-13)}, pages 1067--1075, 2013.

\bibitem{Yang2015AKernels}
Zichao Yang, Andrew Wilson, Alex Smola, and Le~Song.
\newblock {A la Carte – Learning Fast Kernels}, 2015.

\bibitem{Yaglom1987CorrelationFunctions}
A.~M. Yaglom.
\newblock {\em {Correlation Theory of Stationary and Related Random
  Functions}}.
\newblock Springer Series in Statistics. Springer New York, New York, NY, 1987.

\bibitem{Remes2017Non-StationaryKernels}
Sami Remes, Markus Heinonen, and Samuel Kaski.
\newblock {Non-Stationary Spectral Kernels}.
\newblock {\em [Accepted to NIPS 2017] arXiv preprint arXiv:1705.08736}, 2017.

\bibitem{Srivastava2014}
Nitish Srivastava, Geoffrey Hinton, Alex Krizhevsky, Ilya Sutskever, and Ruslan
  Salakhutdinov.
\newblock {Dropout: A Simple Way to Prevent Neural Networks from Overfitting}.
\newblock {\em Journal of Machine Learning Research}, 15:1929--1958, 2014.

\bibitem{Bhatt2017}
Samir Bhatt, Ewan Cameron, Seth~R Flaxman, Daniel~J Weiss, David~L Smith, and
  Peter~W Gething.
\newblock {Improved prediction accuracy for disease risk mapping using Gaussian
  process stacked generalization.}
\newblock {\em Journal of the Royal Society, Interface}, 14(134):20170520, sep
  2017.

\bibitem{Hastie2009TheLearning}
Trevor Hastie, Robert Tibshirani, and J~H Friedman.
\newblock {\em {The elements of statistical learning}}.
\newblock Springer, 2009.

\bibitem{Finkenstadt2007StatisticalSystems}
Barbel. Finkenstadt, Leonhard. Held, and Valerie. Isham.
\newblock {\em {Statistical methods for spatio-temporal systems}}.
\newblock Chapman {\&} Hall/CRC, 2007.

\bibitem{Genest1986TheMarginals}
Christian Genest and Jock MacKay.
\newblock {The Joy of Copulas: Bivariate Distributions with Uniform Marginals}.
\newblock {\em The American Statistician}, 40(4):280, 11 1986.

\bibitem{Gal2015ImprovingInputs}
Yarin Gal and Richard Turner.
\newblock {Improving the Gaussian Process Sparse Spectrum Approximation by
  Representing Uncertainty in Frequency Inputs}.
\newblock {\em Proceedings of the 32nd International Conference on Machine
  Learning (ICML-15)}, 2015.

\bibitem{Tan2013VariationalRegression}
Linda S~L Tan, Victor M~H Ong, David~J Nott, and Ajay Jasra.
\newblock {Variational inference for sparse spectrum Gaussian process
  regression}.
\newblock {\em arXiv preprint arXiv:1306.1999}, 2013.

\bibitem{Baldi2013UnderstandingDropout}
Pierre Baldi and Peter~J Sadowski.
\newblock {Understanding Dropout}, 2013.

\bibitem{Lee2017First-orderPoints}
Jason~D Lee, Ioannis Panageas, Georgios Piliouras, Max Simchowitz, Michael~I
  Jordan, and Benjamin Recht.
\newblock {First-order Methods Almost Always Avoid Saddle Points}.
\newblock {\em arXiv preprint arXiv:1710.07406}, 2017.

\bibitem{Kingma2014Adam:Optimization}
Diederik~P Kingma and Jimmy Ba.
\newblock {Adam: A Method for Stochastic Optimization}.
\newblock {\em arXiv preprint arXiv:1412.6980}, 2014.

\bibitem{Abadi2016TensorFlow:Systems}
Martín Abadi, Ashish Agarwal, Paul Barham, Eugene Brevdo, Zhifeng Chen, Craig
  Citro, Greg~S. Corrado, Andy Davis, Jeffrey Dean, Matthieu Devin, Sanjay
  Ghemawat, Ian Goodfellow, Andrew Harp, Geoffrey Irving, Michael Isard,
  Yangqing Jia, Rafal Jozefowicz, Lukasz Kaiser, Manjunath Kudlur, Josh
  Levenberg, Dan Mane, Rajat Monga, Sherry Moore, Derek Murray, Chris Olah,
  Mike Schuster, Jonathon Shlens, Benoit Steiner, Ilya Sutskever, Kunal Talwar,
  Paul Tucker, Vincent Vanhoucke, Vijay Vasudevan, Fernanda Viegas, Oriol
  Vinyals, Pete Warden, Martin Wattenberg, Martin Wicke, Yuan Yu, and Xiaoqiang
  Zheng.
\newblock {TensorFlow: Large-Scale Machine Learning on Heterogeneous
  Distributed Systems}.
\newblock 3 2016.

\bibitem{Prechelt1998EarlyWhen}
Lutz Prechelt.
\newblock {Early Stopping - But When?}
\newblock pages 55--69. Springer, Berlin, Heidelberg, 1998.

\bibitem{Wan2002ValidationData}
Zhengming Wan, Yulin Zhang, Qincheng Zhang, and Zhao-liang Li.
\newblock {Validation of the land-surface temperature products retrieved from
  Terra Moderate Resolution Imaging Spectroradiometer data}.
\newblock {\em Remote Sensing of Environment}, 83(1-2):163--180, 11 2002.

\bibitem{Wu2007OnSeries.}
Zhaohua Wu, Norden~E Huang, Steven~R Long, and Chung-Kang Peng.
\newblock {On the trend, detrending, and variability of nonlinear and
  nonstationary time series.}
\newblock {\em Proceedings of the National Academy of Sciences of the United
  States of America}, 104(38):14889--94, 9 2007.

\bibitem{LeCun2015DeepLearning}
Yann LeCun, Yoshua Bengio, and Geoffrey Hinton.
\newblock {Deep learning}.
\newblock {\em Nature}, 521(7553):436--444, 5 2015.

\bibitem{2014TrendsAnalytics}
{Trends in big data analytics}.
\newblock {\em Journal of Parallel and Distributed Computing},
  74(7):2561--2573, 7 2014.

\bibitem{Rue2009ApproximateApproximationsb}
H{\aa}vard Rue, Sara Martino, and Nicolas Chopin.
\newblock {Approximate Bayesian inference for latent Gaussian models by using
  integrated nested Laplace approximations}.
\newblock {\em Journal of the Royal Statistical Society: Series B (Statistical
  Methodology)}, 71(2):319--392, 2009.

\bibitem{Minka2001ExpectationInference}
Thomas~P Minka.
\newblock {Expectation propagation for approximate Bayesian inference}.
\newblock pages 362--369, 2001.

\bibitem{CarpenterStan:Language}
Bob Carpenter, Bob Carpenter, Daniel Lee, Marcus~A. Brubaker, Allen Riddell,
  Andrew Gelman, Ben Goodrich, Jiqiang Guo, Matt Hoffman, Michael Betancourt,
  and Peter Li.
\newblock {Stan: A Probabilistic Programming Language}.

\bibitem{Bengio2012PracticalArchitectures}
Yoshua Bengio.
\newblock {Practical Recommendations for Gradient-Based Training of Deep
  Architectures}.
\newblock pages 437--478. Springer Berlin Heidelberg, 2012.

\bibitem{Chen2014StochasticCarlo}
Tianqi Chen, Emily Fox, and Carlos Guestrin.
\newblock {Stochastic Gradient Hamiltonian Monte Carlo}, 1 2014.

\bibitem{Lee2017DeepProcesses}
Jaehoon Lee, Yasaman Bahri, Roman Novak, Samuel~S. Schoenholz, Jeffrey
  Pennington, and Jascha Sohl-Dickstein.
\newblock {Deep Neural Networks as Gaussian Processes}.
\newblock 10 2017.

\end{thebibliography}
\end{document}